\documentclass[runningheads]{llncs}

\usepackage[T1]{fontenc}
\usepackage{graphicx}
\usepackage{color}
\usepackage[misc]{ifsym}
\usepackage{bbding}
\usepackage{multirow}
\usepackage{subfiles}
\usepackage{hyperref}
\usepackage{subcaption}
\usepackage{longtable}
\usepackage{cite}
\usepackage{colortbl}
\usepackage{booktabs}
\usepackage{multirow}
\usepackage{amsmath}

\usepackage[table]{xcolor}
\definecolor{maroon}{cmyk}{0,0.87,0.68,0.32}

\newcolumntype{L}[1]{>{\raggedright\arraybackslash\hspace{#1}}p{#1}}
\newcolumntype{P}[1]{>{\raggedright\arraybackslash}p{#1}} 

\begin{document}

\title{Deep Neural Networks for Predicting Recurrence and Survival in Patients with Esophageal Cancer After Surgery}

\titlerunning{Deep Neural Networks for Esophageal Cancer Prognosis Prediction}

\author{Yuhan Zheng\inst{1}$^($\Envelope$^)$\orcidID{0000-0002-9762-6530} \and Jessie A Elliott \inst{2} \and John V Reynolds \inst{2} \and Sheraz R Markar \inst{3} \and Bart\l omiej W. Papie\.{z}\inst{1}$^($\Envelope$^)$\orcidID{0000-0002-8432-2511} \and ENSURE study group \inst{4}}

\institute{Big Data Institute, University of Oxford, Oxford, United Kingdom \\ \email{yuhan.zheng@univ.ox.ac.uk, bartlomiej.papiez@bdi.ox.ac.uk} \and
Trinity St. James's Cancer Institute, Trinity College Dublin and St. James's Hospital, Dublin, Ireland \and  Nuffield Department of Surgery, University of Oxford, United Kingdom \and Young Investigator Division, European Society for Diseases of the Esophagus}

\authorrunning{Y. Zheng et al.}

\maketitle             

\begin{abstract}
Esophageal cancer is a major cause of cancer-related mortality internationally, with high recurrence rates and poor survival even among patients treated with curative-intent surgery. Investigating relevant prognostic factors and predicting prognosis can enhance post-operative clinical decision-making and potentially improve patients' outcomes. In this work, we assessed prognostic factor identification and discriminative performances of three models for Disease-Free Survival (DFS) and Overall Survival (OS) using a large multicenter international dataset from ENSURE study. We first employed Cox Proportional Hazards (CoxPH) model to assess the impact of each feature on outcomes. Subsequently, we utilised CoxPH and two deep neural network (DNN)-based models, DeepSurv and DeepHit, to predict DFS and OS. The significant prognostic factors identified by our models were consistent with clinical literature, with post-operative pathologic features showing higher significance than clinical stage features. DeepSurv and DeepHit demonstrated comparable discriminative accuracy to CoxPH, with DeepSurv slightly outperforming in both DFS and OS prediction tasks, achieving C-index of 0.735 and 0.74, respectively. While these results suggested the potential of DNNs as prognostic tools for improving predictive accuracy and providing personalised guidance with respect to risk stratification, CoxPH still remains an adequately good prediction model, with the data used in this study.

\end{abstract}

\section{Introduction}
\label{introduction}

Esophageal cancer is a major cause of cancer-related mortality internationally. The average 5-year Overall Survival (OS) rate is less than 25\%\cite{rebecca}, ranging from 10\% to 55\% depending on the stage of which the disease is detected\cite{fabre}. While surgical resection, known as esophagectomy, remains the primary treatment for esophageal cancer, the prognosis of post-operative patients remains poor. Despite advancements in cancer management strategy, more than 50\% of the patients experience a recurrence within 1-3 years following curative-intent surgery\cite{thomas}, with a median survival time of 24 months\cite{kunisaki}. Therefore, identifying prognostic factors associated with a higher risk of recurrence, as well as predicting and stratifying patients based on their recurrence and survival probabilities, are crucial to the delivery of personalised medicine approaches that could potentially improve oncologic outcomes. Current risk stratification methods for patients with esophageal cancer predominantly rely on pathological data, primarily tumor staging\cite{barbar}. This does not fully leverage all available clinical and patient-level data efficiently, and does not account for individual variations. 


To address these issues, some studies have developed models for prognosis prediction. For example, logistic regression models have been employed to predict absolute risks for patients with esophageal cancer\cite{wang, chen}. However, these models predict a single-point outcome event without incorporating time-to-event analysis and are limited to one histologic type only. The Cox Proportional Hazards model (CoxPH)\cite{cox} is a widely used regression model that allows the study of the relationships between time-to-event outcomes and a set of covariates. Many studies have employed CoxPH to identify prognostic factors for different outcomes\cite{3, 4, 5}. However, CoxPH model assumes linear relationships between covariates and that the relative hazard remains constant over time. This hinders its ability to capture higher level interactions between variables and outcomes. 

Recent developments in AI have led to increased applications of machine learning (ML) models in oncology to address more complex problems. For example, Zhang et al.\cite{zhang} explored multiple ML methods for survival prediction in squamous cell carcinoma, and demonstrated that while CoxPH model remains sufficiently good for interpretive studies, ML approaches have the potential to enhance predictive accuracy. Gong et al.\cite{gong} explored artificial neural networks (ANNs) in survival prediction, though these did not outperform other traditional ML models such as XGBoost\cite{xgboost}. However, most of these aforementioned studies relied on data collected from a single center, raising questions about their generalisability and robustness when applied to larger multicenter cohorts. Most studies focus on only one type of outcome, and the prediction values on other outcomes remain unknown. Moreover, these studies often utilise a limited number of features. There is a significant clinical interest in incorporating a more comprehensive set of features that take account into, for example, improvements in treatment technologies or surveillance strategies. Gujjuri et al.\cite{rohan} implemented CoxPH and Random Forest using ENSURE dataset. However, the results showed that Random Forest did not surpass CoxPH in both discrimination and calibration.

In this work, we developed models to predict Disease-Free Survival (DFS) and OS for patients with esophageal cancer following curative-intent surgery. The work is divided into two main components. The first component is prognostic factor identification task, which aims to identify significant prognostic factors that influence outcomes based on their hazard ratios and significance values, thereby providing clinical guidance. The second component is a prediction task, which aims to develop robust models for prognosis prediction on multicenter heterogeneous dataset. This helps stratify patients based on their risks, which could potentially facilitate personalisation of postoperative treatment and surveillance strategies.

Our contributions are threefold. Firstly, we developed models using a large heterogeneous multicenter cohort\cite{jessie}. Secondly, we incorporated a comprehensive set of easily accessible and readily identifiable features into the models, including several general center-specific features, to explore more broadly prognostic factors. Finally we carried out extensive experiments with deep neural network (DNN)-based models, and compared their predictive performance with CoxPH model.

The remainder of this paper is organised as follows. Section \ref{dataset and preprocessing} introduces the details of the dataset, preprocessing steps and provides an overview of the final dataset used for this work. The three models employed and the experimental setup which includes training and implementation details, are described in Section \ref{models} and Section \ref{experimental setup}, respectively. Section \ref{prognostic factor identification task} and Section \ref{prediction task} present results for the prognostic factor identification task and prediction task. Finally, the discussion and conclusion can be found in Section \ref{discussions and conclusions}.

\section{Dataset and Preprocessing}
\label{dataset and preprocessing}
\subsection{Dataset} 
\label{dataset}

This work is based on data collected from the European iNvestigation of SUrveillance after Resection for Esophageal cancer (ENSURE) study\cite{jessie}, a retrospective non-interventional study taken across 20 European centers. Patients with esophageal or junction cancer undergoing curative intent treatment from June 2009 to June 2015 were all considered for inclusion. In total, there are 4972 patients and over 170 variables. All patients were staged according to the 8th edition of the American Joint Committee on Cancer (AJCC) staging\cite{ajc}.

The use of the dataset and this study has been approved by he Joint Research Ethics Committee of Tallaght University Hospital and St. James’s Hospital, Dublin, Ireland (SJH-TUH JREC Ref 2943 Amendment 1).

\subsection{Outcome Variable Definition}
\label{outcome variable definition}

In this work, DFS is defined as the time from treatment (i.e., surgery) to recurrence or death from any cause\cite{si}. Patients who are lost to follow-up or remain alive without recurrence at the end of the study are recorded as censored events. OS is defined as the period from diagnosis to death from any cause\cite{david}. Patients that are lost to follow-up or still alive at the end of the study are recorded as censored event. 

\subsection{Patient Inclusion and Variable Selection Criteria}
\label{patient inclusion and variable selection criteria}

\subsubsection{Patient Inclusion.}

In this work, we removed patients with missing DFS and/or OS outcome, as well as patients with rare histologic type (i.e., non-adenocarcinoma and non-squamous cell carcinoma). We excluded further patients with postoperative death for DFS prediction by definition.

\subsubsection{Variable Selection.}

Variables used in our models were selected by experienced clinicians, based on the literature review and their clinical importance. Variables exhibiting clinically known high correlations with other variables, lacking well-established relationships with outcomes, or variables that were often poorly documented by centers, were excluded from the study. Additionally, while there is no single acceptable threshold for missing rate, the approach to dealing with missingness requires careful consideration. Blindly applying imputation strategies to variables with high missing rate could also impose biases\cite{dong}. Therefore, after further assessment by clinicians, a set of variables was additionally removed based on both their rate of missingness and their clinical relevance. 

In this work, we did not apply any ML or statistic-based variable selection strategies. Evidence\cite{spooner} suggests that feature selection prior to model application does not significantly improve model performance, especially that we either adopted regularisers in the model (more details in Section \ref{experimental setup}) or the ML models themselves have internal feature selection capabilities to handle high-dimensional data in this study. As a result of this variable selection processes, 37 variables were selected with a missing rate of less than 30\%.

\subsection{Missingness and Imputation}
\label{missingness and imputation}

In this study, the missingness mechanism was assumed to be Missing At Random (MAR)\cite{mack}, as whether the data is missing or not depends exclusively on their availability at center during data collection process\cite{pedro, jerez}. This assumption allows us to apply imputation strategies to handle missingness. A flow chart illustrating the overall process, which is going to described below, can be found in Figure \ref{figure3} in Section \ref{imputation procedure}.

Different imputation strategies were applied to the prognostic factor identification task and prediction task that were mentioned in Section \ref{introduction}. Multiple Imputation by Chained Equations (MICE)\cite{azur} was used for prognostic factor identification task, with 10 iterations per imputation set. Multiple imputation (MI), which takes the uncertainty of imputation into account and fills different multiple plausible values, is important to reduce bias and chance of false-positive and false-negative conclusions\cite{li}. The multiple imputed datasets were passed into models, optimised and analysed separately, and final results were combined using Rubin's rule\cite{rubin}. For prediction task, where the impact of imputation uncertainty is generally less critical, we used single-point multivariate imputation by chained equations, which is typically sufficient for predictive modeling purposes. 

When performing imputation, outcome variables, including the binary event indicator and the time-to-event variable, were also included in the prediction matrix to prevent bias\cite{austin}. The time-to-event variable was transformed to its cumulative hazard function with the non-parametric Nelson-Aalen estimator\cite{enrico} as suggested in \cite{white}. The imputation was conducted within the cross-validation (CV) loop during training\cite{spooner} to prevent any information leakage from the validation set into the training process. 

Prior to imputation, the nominal categorical variables were dummy coded. It is important to note that during the imputation process, continuous values were generated for all dummy-coded binary variables, and these values were not rounded to the nearest integer, as recommended based on the findings in \cite{ake}. Additionally, after imputation, continuous numerical variables were scaled by zero-score standardisation to bring all variables to approximately similar dynamic ranges to improve numerical stability during training. 

\begin{table}[h]
\caption{Summary of the dataset used for model development. DFS task (n=3921), OS task (n=4077).}
\vspace{3mm}
\centering
\begin{tabular}{c|c|c|c|c|c|c|c}
\toprule
\multirow{2}{*}{Outcome} & \multirow{2}{*}{\shortstack{No. of\\Variables}} & \multirow{2}{*}{\shortstack{No. of\\Patients}} & \multirow{2}{*}{\shortstack{No. of\\Observed Events}} & \multirow{2}{*}{\shortstack{Min.\\(months)}} & \multirow{2}{*}{\shortstack{Max.\\(months)}} & \multirow{2}{*}{\shortstack{Median\\(months)}} & \multirow{2}{*}{\shortstack{Mean\\(months)}} \\ 
& & & & & & & \\
\midrule
DFS & 34 & 3921 & 2308 & 0 & 173 & 29.7 & 36.1 \\
OS & 34 & 4077 & 2173 & 0.2 & 176.7 & 37.47 & 41.92 \\
\bottomrule
\end{tabular}
\label{table1}
\end{table}

Furthermore, after standardisation, three variables that had Pearson correlation coefficients higher than 70\% were removed. While there is no definitive threshold for exclusion, we set this threshold based on the interpretations provided in \cite{akoglu} and common practices in the field. As a result, 34 variables were ultimately selected for model development.

\subsection{Data Overview}
\label{data overview}

Table \ref{table1} summarises the statistics for the dataset used in DFS and OS tasks, respectively.

\section{Methods and Experiments}
\label{methods and experiments}

\subsection{Models}
\label{models}

In this work, three models were employed to predict DFS and OS: a regression model CoxPH\cite{cox} and two neural network-based models named DeepSurv\cite{deepsurv} and DeepHit\cite{deephit}. CoxPH is a semi-parametric regression model that takes the form $h_0(t)exp(\sum_i x_i \cdot \beta_i)$, where $h_0(t)$ is baseline hazard function, $x_i$ is covariate and $\beta_i$ is coefficient. The model assumes that the effect of a factor is constant over time and there is a linear relationship between predictors and log-hazards. DeepSurv is a DNN-based extension of CoxPH model. It models the hazard function as $h_0(t)exp(f_{\boldsymbol{\theta}}(\boldsymbol{x}))$, where $f_{\boldsymbol{\theta}}(\boldsymbol{x})$ is a neural network that takes covariates as input and outputs a scalar. This allows DeepSurv to capture high-level interactions among features. DeepHit, on the other hand, employs an end-to-end DNN that learns the distribution of survival times directly, without making any assumptions about the underlying stochastic process.

In this work, CoxPH model was employed for the prognostic factor identification task. For the prediction task, all three models were used, with CoxPH serving as a baseline for comparison with neural network-based methods. These models were chosen to leverage their respective strengths in handling different aspects of survival analysis, from traditional regression assumptions to capturing complex interactions and learning distributions directly from data.

\subsection{Experimental Setup}
\label{experimental setup}
\subsubsection{Dataset Splitting Strategy.} 

The dataset was split into two parts: 80\% for training and 20\% as held-out testing dataset. For the prognostic factor identification task, the training set was further split into 85\% for training and 15\% for validation. Stratified bootstrapping was performed on the validation set to select the best set of hyperparameters. For the prediction task, a stratified 5-fold CV was performed on the 80\% training set for hyperparameter selection. The imputation and standardisation were performed within the CV loop to avoid information leakage, as mentioned in Section \ref{missingness and imputation}. A graphical illustration of the splitting strategy can be found in Figure \ref{figure2} in Appendix Section \ref{splitting strategy}. 

\subsubsection{Hyperparameter Tuning.}

Hyperparameter selection was conducted in a grid-search manner. A detailed list of the optimal set of hyperparameters for each model and task can be found in Table \ref{tables1} in Appendix Section \ref{appendix: hyperparameter tuning}. Elastic net regularisation (i.e., L1 (Lasso) and L2 (Ridge) regularisation penalties) was applied to CoxPH. 
CoxPH with Elastic net \cite{elasticnet} was generally found to outperform standard CoxPH during training. 

\subsubsection{Performance Evaluation.}

Three metrics were used to evaluate the discrminative performances of the models: concordance index (C-index), Integrated Brier Score (IBS), and time-dependent AUC (tAUC, also known as dynamic AUC). 

\subsubsection{Implementation.}

All the models and analyses were implemented using Python 3.10.5. Survival models were implemented with lifelines 0.28.0 and pycox 0.2.3. The CoxPH was trained on a CPU with a memory of 15.2GB. DeepSurv and DeepHit were trained on NVIDIA GPUs with 40GB of RAM.

\begin{table}[!htbp]
\centering
\caption{Multivariate CoxPH analysis results for DFS and OS. Relative hazard ratio was calculated for nominal categorical variables with one category as reference (indicated as `ref' in the table). P < 0.05 was considered as significant. Only significant variables are listed here. NA: neoadjuvant; CRT: chemoradiation therapy. Definitions and staging criteria of the features can be found in \cite{ajc}.}
\label{table2}
\begin{tabular}{@{}p{3.47cm}>
{\centering\arraybackslash}p{2.96cm}>{\centering\arraybackslash}p{1.23cm}>{\centering\arraybackslash}p{2.96cm}>{\centering\arraybackslash}p{1.23cm}@{}}
\toprule
 & \multicolumn{2}{c}{\textbf{DFS}} & \multicolumn{2}{c}{\textbf{OS}} \\
\cmidrule(lr){2-3} \cmidrule(lr){4-5}
\textbf{Variable} & \textbf{HR (95\% CI)} & \textbf{P-value} & \textbf{HR (95\% CI)} & \textbf{P-value}\\
\midrule

\textbf{Sex} & & & &  \\
\hspace{1em}Female & ref & & & \\
\hspace{1em}Male & 1.200 (1.199-1.200) & \textbf{0.007} & 1.134 (1.130-1.138) & 0.050 \\

\textbf{Age (Years)} & 1.017 (1.001-1.033) & 0.553 & 1.115 (1.114-1.115) & \textbf{<0.001} \\

\textbf{Clinical N stage} & & & &  \\
\hspace{1em}cN0 & ref & & & \\
\hspace{1em}cN2 & 1.155 (1.147-1.163) & 0.081 & 1.239 (1.238-1.240) & \textbf{0.010} \\

\textbf{Tumor Site} & & & &  \\
\hspace{1em}Junctional & ref & & & \\
\hspace{1em}Lower & 1.207 (1.206-1.207) & \textbf{0.006} & 1.135 (1.132-1.138) & \textbf{0.042} \\
\hspace{1em}Middle & 1.309 (1.306-1.311) & \textbf{0.019} & 1.266 (1.262-1.269) & \textbf{0.027} \\

\textbf{Proximal margin positive} & 1.994 (1.994-1.994) & \textbf{<0.001} & 1.292 (1.267-1.318) & 0.121 \\

\textbf{Radial margin positive} & 1.549 (1.549-1.549) & \textbf{<0.001}& 1.426 (1.426-1.426) & \textbf{<0.001} \\

\textbf{Pathologic T stage} & & & &  \\
\hspace{1em}T0 & ref & & &  \\
\hspace{1em}T3 & 1.583 (1.583-1.583) & \textbf{<0.001} & 1.490 (1.490-1.490) & \textbf{0.001} \\
\hspace{1em}T4 & 1.672 (1.671-1.672) & \textbf{0.002} & 1.752 (1.751-1.752) & \textbf{0.002} \\

\textbf{Pathologic N stage} & & & &  \\
\hspace{1em}N0 & ref & & & \\
\hspace{1em}N1 & 1.484 (1.484-1.484) & \textbf{<0.001} & 1.245 (1.243-1.246) & \textbf{0.013} \\
\hspace{1em}N2 & 1.664 (1.664-1.664) & \textbf{<0.001} & 1.528 (1.528-1.528) & \textbf{<0.001} \\
\hspace{1em}N3 & 3.087 (3.087-3.087) & \textbf{<0.001} & 2.991 (2.991-2.991) & \textbf{<0.001} \\

\textbf{Pathologic M stage} & & & &  \\
\hspace{1em}M0 & ref & & & \\
\hspace{1em}M1 & 1.707 (1.707-1.707) & \textbf{<0.001} & 1.919 (1.919-1.919) & \textbf{<0.001} \\

\textbf{Differentiation} & & & &  \\
\hspace{1em}Gx, cannot be assessed  & ref & & &  \\
\hspace{1em}Poorly differentiated  & 1.379 (1.378-1.379) & \textbf{0.004} & 1.447 (1.446-1.447) & \textbf{0.003} \\

\textbf{Lymphatic invasion} & 1.055 (1.000-1.113) & 0.573 & 1.316 (1.316-1.316) & \textbf{<0.001} \\

\textbf{Venous invasion} & 1.292 (1.291-1.292) & \textbf{<0.001} & 1.086 (1.060-1.112) & 0.306 \\

\textbf{Perineural invasion} & 1.161 (1.158-1.164) & \textbf{0.038} & 1.193 (1.193, 1.194) & \textbf{0.006} \\

\textbf{Number of nodes analyzed} & 0.894 (0.894, 0.894) & \textbf{0.002} & 0.883 (0.884, 0.884)& \textbf{<0.001}\\

\textbf{Treatment protocol} & & & &  \\
\hspace{1em}Surgery only  & ref & & &  \\
\hspace{1em}NA CRT then surgery & 1.326 (1.326-1.326) & \textbf{0.003} & 1.198 (1.195-1.200) & \textbf{0.025} \\

\textbf{Clavien-Dindo Grade} & 1.060 (1.059-1.060) & \textbf{<0.001} & 1.169 (1.169-1.169) & \textbf{<0.001} \\
\textbf{Length of stay (Days)} & 1.077 (1.076-1.077) & \textbf{0.010} & 1.052 (1.050-1.053) & \textbf{0.044} \\
\textbf{Cancer cases per year} & 0.919 (0.919-0.919) & \textbf{0.008} & 0.925 (0.924-0.926) & \textbf{0.024} \\

\bottomrule

\end{tabular}
\end{table}

\section{Results}
\label{results}

\subsection{Prognostic Factor Identification Task}
\label{prognostic factor identification task}

Table \ref{table2} summarises the multivariate analysis results of CoxPH with the significant variables (P-value < 0.05) being listed only, along with their hazard ratio (HR), and 95\% confidence interval (CI).

\subsection{Prediction Task}
\label{prediction task}

Table \ref{table3} summarises the discriminative prediction performance of three models for DFS and OS respectively. Comparing all three metrics reveals that DeepSurv demonstrates comparable performances to CoxPH, while DeepHit demonstrates slightly inferior performance in terms of IBS. Figure \ref{figure1} in Appendix Section \ref{appendix: predicted survival curves} provides examples of predicted OS curves obtained from the three models for the same random set of five patients. Notably, while CoxPH and DeepSurv exhibit similar shapes and distributions, DeepHit shows a completely different profile, with minimal variation among the five prediction curves. Despite DeepHit generally ordering patients consistently in terms of survival probabilities compared to the other two models, this profile suggests poorer calibration performance.

\begin{table}[h]
\caption{Summary of model performances. C-index: concordance index; IBS: Integrated Brier Score; tAUC: time-dependent AUC.}
\vspace{3mm}
\centering
\begin{tabular}{c|c|c|c}

\hline
 & C-index (95\% CI) $\uparrow$ & IBS (95\% CI) $\downarrow$ & tAUC (95\% CI) $\uparrow$ \\
\hline
\textbf{DFS} &  &  &    \\
CoxPH & 0.733 (0.710, 0.755) & \textbf{0.174 (0.160, 0.187)} & 0.720 (0.682, 0.799)\\
DeepSurv & \textbf{0.735 (0.714, 0.758)} & 0.176 (0.163, 0.193) & \textbf{0.749 (0.727, 0.801)} \\
DeepHit & 0.729 (0.707, 0.752) & 0.249 (0.243, 0.263) & 0.729(0.693, 0.797) \\
\textbf{OS} & &  &    \\
CoxPH & 0.734 (0.710, 0.758) & \textbf{0.164 (0.153, 0.181)} & \textbf{0.783 (0.738, 0.818)} \\
DeepSurv & \textbf{0.740 (0.716, 0.764)} & 0.169 (0.152, 0.192) & 0.781 (0.734, 0.827) \\
DeepHit & 0.739 (0.716, 0.762)& 0.214 (0.201, 0.233) & 0.776 (0.707, 0.827) \\

\hline
\end{tabular}
\label{table3}
\end{table}

\section{Discussion and Conclusion}
\label{discussions and conclusions}

In this work, we analysed a heterogeneous multicenter dataset to investigate the contribution of covariates to and predictive performance of three models on DFS and OS in patients with esophageal cancer. 
The significant prognostic factors identified aligned well with clinical literature and experiences. For example, pathologic tumor staging features appear to be strong prognostic factors, and are generally more significant than clinical staging\cite{1}. The more advanced the pathologic stage of the tumor is, the higher the hazard ratio. In terms of prediction, DeepSurv consistently outperformed CoxPH in both DFS and OS tasks, with C-index of 0.735 and 0.740, respectively, when C-index serving as the primary metric. Overall, the two DNN-based models demonstrated comparable discriminative performance to CoxPH; though DeepHit was found to exhibit poorer calibration performance compared to the other two models. The use of a multicenter international dataset, which includes patients with either adenocarcinoma or squamous cell carcinoma, suggested broader applicability of these findings across diverse cohort in various clinical settings. 
In general, despite their ability to model more complex interactions, DNN-based models did not greatly outperform the CoxPH. The CoxPH, which is interpretable and computationally efficient, still remains a sufficiently good prediction model with tabular data. 

While all three models demonstrated good discriminative performance, it is inferred that these results likely represent the upper bound achievable with tabular data. It is worth noting that some significant features, for example, Clinical N stage, are derived from radiologic assessment scans (Computed Tomography (CT), Positron Emission Tomography (PET))\cite{ajc}. Therefore, incorporating imaging-derived features such as radiomics could provide more detailed information and potentially enhance model performance\cite{mona}. It should be noted that, among the three models, DeepHit posed particular challenges during training, showing large fluctuations in performance and high sensitivity to hyperparameters. This difficulty can be attributed to its end-to-end neural network architecture, which involves a multitude of hyperparameters. More advanced hyperparameter selection techniques such as Bayesian optimisation could therefore be explored\cite{hyperopti1} during the training. Graphical convolutional neural work (GNN)\cite{graph1}, which has been an emerging model in survival analysis, could also be explored in the future.
In addition, it could be observed that DFS and OS share some common significant prognostic factors. This suggest the possibility of multi-task learning of these two prediction tasks\cite{multi1}. 
Furthermore, introducing additional calibration techniques\cite{caliloss1} could improve the alignment of predictions with ground truth data. Methods like SHAP \cite{shap} could also enhance the interpretability of neural networks by identifying crucial features in predictive models.

\begin{credits}

\subsubsection{\ackname} This work is funded by Cancer Research UK (CRUK). The authors acknowledge the contributions of Sinead King, St. James’s Hospital, Dublin, Ireland; Hannah Adams, Gloucestershire Hospitals NHS Foundation Trust, England; and Masaru Hayami, CLINTEC, Karolinska Institutet, Stockholm, Sweden. The authors acknowledge the contributions to and previous works on the ENSURE study by Elliott J.A. et al. \cite{jessie} and Gujjuri R.R. et al. \cite{rohan}.

The authors would like to thank the Oxford Biomedical Research Computing (BMRC) facility for providing the computing resources. Special thanks to Lav Radosavljevic from University of Oxford for his professional advice on statistical analysis.

\subsubsection{\discintname}
The authors have no competing interests to declare that are
relevant to the content of this article.
\end{credits}

\appendix
\section{Appendix}

\subsection{Imputation Procedure}
\label{imputation procedure}

\begin{figure} [!htb]

  \centering
  \includegraphics[width=\textwidth]{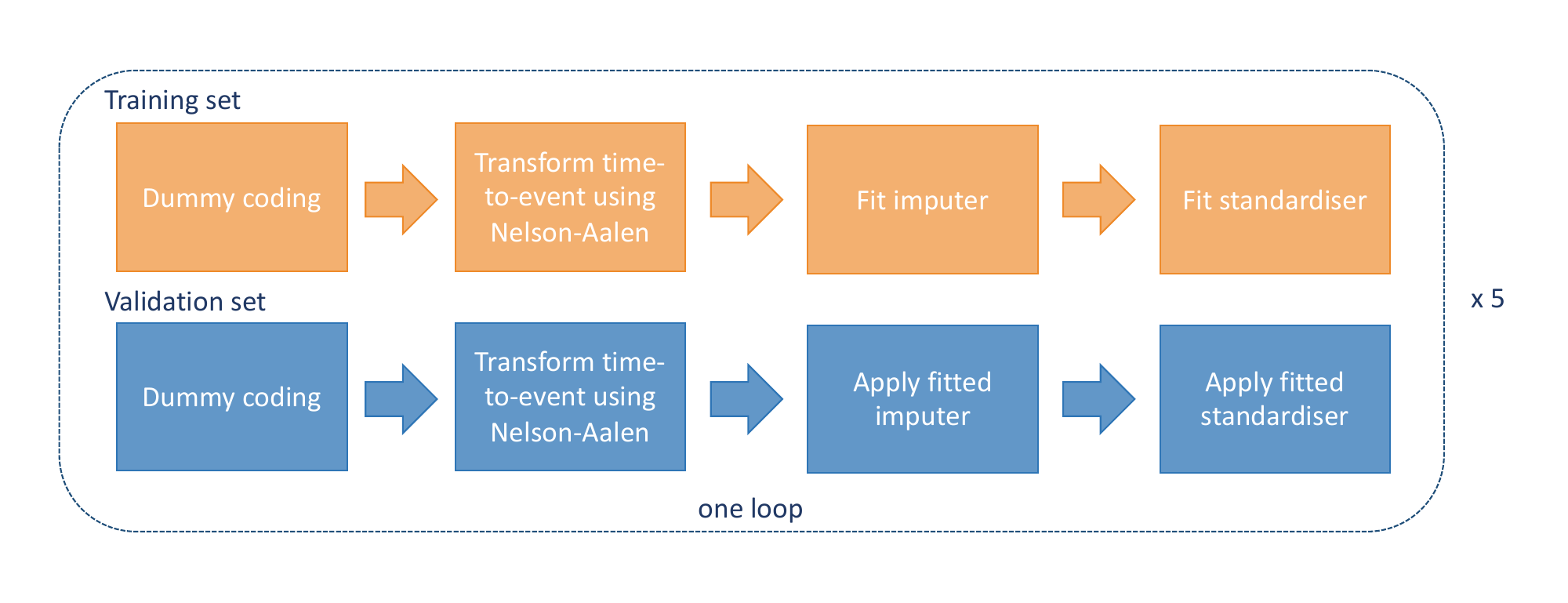}
  \caption{Flowchart illustrating the preprocessing and imputation process. The process is performed for each loop and repeated across all loops within the CV.}
  \label{figure3}

\end{figure}

Figure \ref{figure3} illustrates the overall process of variable preprocessing and imputation, as described in Section \ref{missingness and imputation}.

\subsection{Splitting Strategy}
\label{splitting strategy}

\begin{figure} [!htb]

  \centering
  \includegraphics[width=\textwidth]{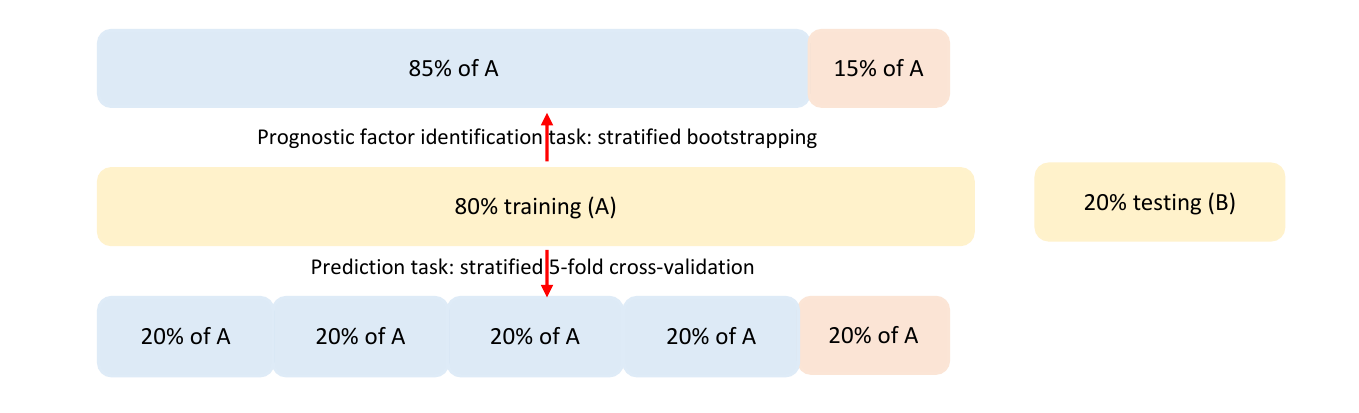}
  \caption{Graphical illustration of the splitting strategy during training for prognostic factor identification task and prediction task, where blue and orange color represents the training and validation set for each task, respectively.}
  \label{figure2}

\end{figure}

Figure \ref{figure2} illustrates the splitting strategy during training for the two tasks.

\subsection{Hyperparameter Tuning}
\label{appendix: hyperparameter tuning}

Table \ref{tables1} summarises the optimal set of hyperparameters of the three models. For DeepSurv and DeepHit, various network structures were explored, along with different number of epochs, batch sizes, optimiser schedulers, and learning rates.

\begin{table}[!htb]
\caption{Summary of the optimal hyperparameter set of the three models for DFS and OS.}
\vspace{3mm}

\begin{subtable}[c]{\linewidth}
\centering
\caption{Best hyperparameter set of CoxPH.}
\begin{tabular}{c|c|c}
\hline
& \textbf{L1 penalty} & \textbf{L2 penalty} \\
\hline
\textbf{DFS} & 0.008 & 0.001 \\
\textbf{OS} & 0.006 & 0.002 \\
\hline
\end{tabular}
\label{tables1a}
\end{subtable}

\begin{subtable}[c]{\linewidth}
\centering
\caption{Best hyperparameter set of DeepHit. lr: learning rate; w decay: weight decay.}
\begin{tabular}{c|c|c|c|c}
\hline
& \textbf{network} & \textbf{dropout} & \textbf{epochs} & \textbf{batch size} \\
\hline
\textbf{DFS} & [64, 64] & 0.1 & 75 & 64 \\
\textbf{OS} & [64, 128, 64] & 0.1 & 70 & 256 \\
\hline
& \textbf{optimiser} & \textbf{initial lr} & \textbf{scheduler} &  \textbf{w decay} \\
\hline
\textbf{DFS} & Adam & 0.1 & Exp.LR, $\gamma$=0.7 & 0.05 \\
\textbf{OS} & Adam & 0.1 & Exp.LR, $\gamma$=0.7 & 0.05 \\
\hline
\end{tabular}
\label{tables1c}
\end{subtable}

\begin{subtable}[c]{\linewidth}
\centering
\caption{Best hyperparameter set of DeepHit. lr: learning rate; w decay: weight decay.}
\begin{tabular}{c|c|c|c|c|c}
\hline
& \textbf{network} & \textbf{dropout} & \textbf{epochs} & \textbf{batch size} & \textbf{optimiser}\\
\hline
\textbf{DFS} & [64, 128, 64] & 0.1 & 25 & 64 & Adam \\
\textbf{OS} & [64, 128, 64] & 0.1 & 100 & 64 & Adam \\
\hline
& \textbf{initial lr} & \textbf{scheduler} &  \textbf{w decay} & \textbf{no. of output} & \textbf{output interp. no.} \\
\hline
\textbf{DFS} & 0.005 & Exp.LR, $\gamma$=0.7 & 0.05 & 60 & 50 \\
\textbf{OS} & 0.005 & Exp.LR, $\gamma$=0.7 & 0.05 & 60 & 50 \\
\hline
\end{tabular}
\label{tables1c}
\end{subtable}

\label{tables1}

\end{table}

\subsection{Predicted Survival Curves}
\label{appendix: predicted survival curves}

\begin{figure}[h]
\centering

\begin{subfigure}{0.45\textwidth}
\centering
\includegraphics[width=\textwidth,height=0.25\textheight]{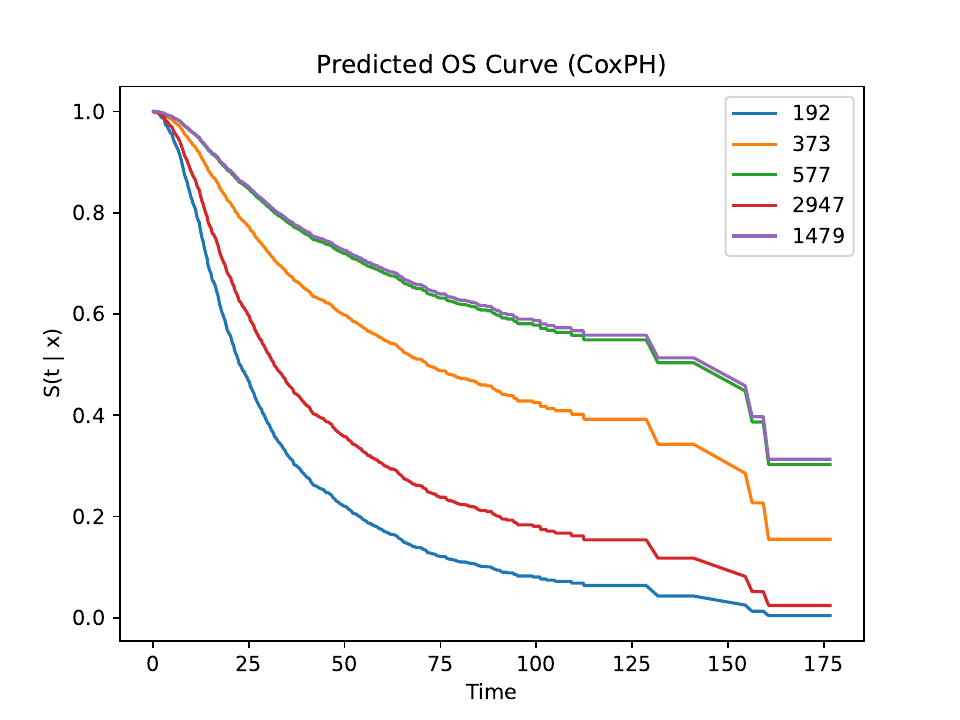}
\caption{CoxPH}
\label{figure1a}
\end{subfigure}
\begin{subfigure}{0.45\textwidth}
\centering
\includegraphics[width=\textwidth,height=0.25\textheight]{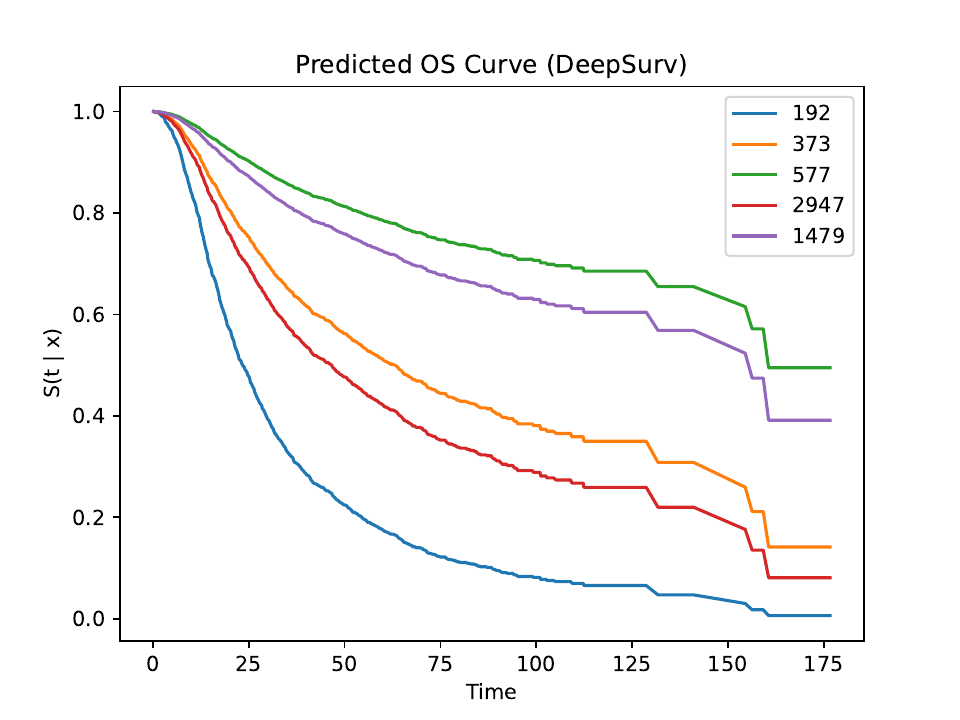}
\caption{DeepSurv}
\label{figure1b}
\end{subfigure}
\begin{subfigure}{0.45\textwidth}
\centering
\includegraphics[width=\textwidth,height=0.25\textheight]{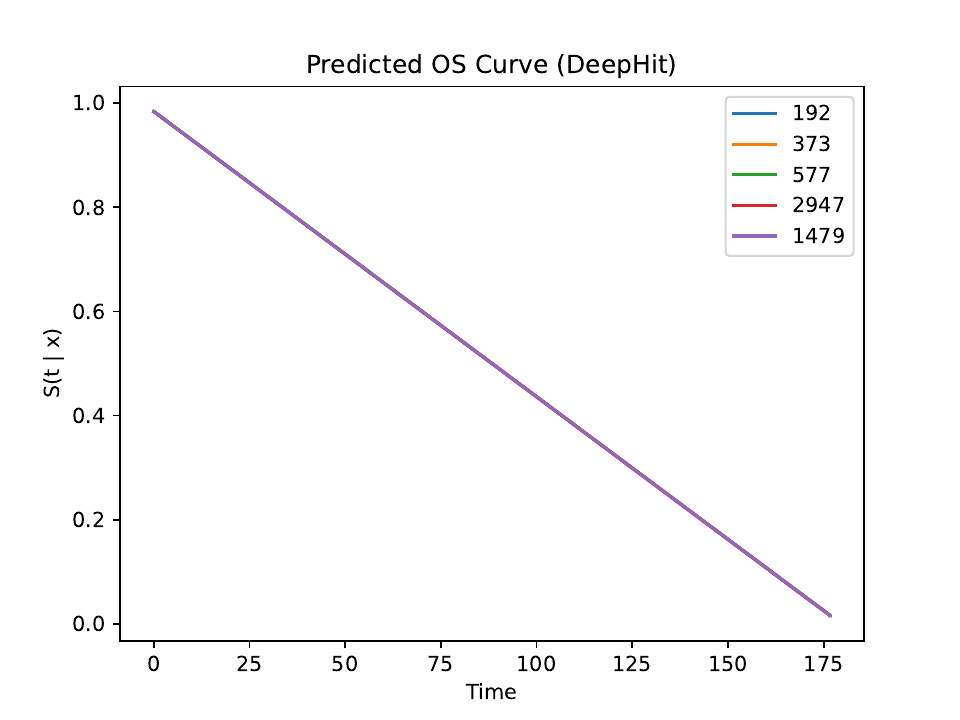}
\caption{DeepHit}
\label{figure1c}
\end{subfigure}
\begin{subfigure}{0.45\textwidth}
\centering
\includegraphics[width=\textwidth,height=0.25\textheight]{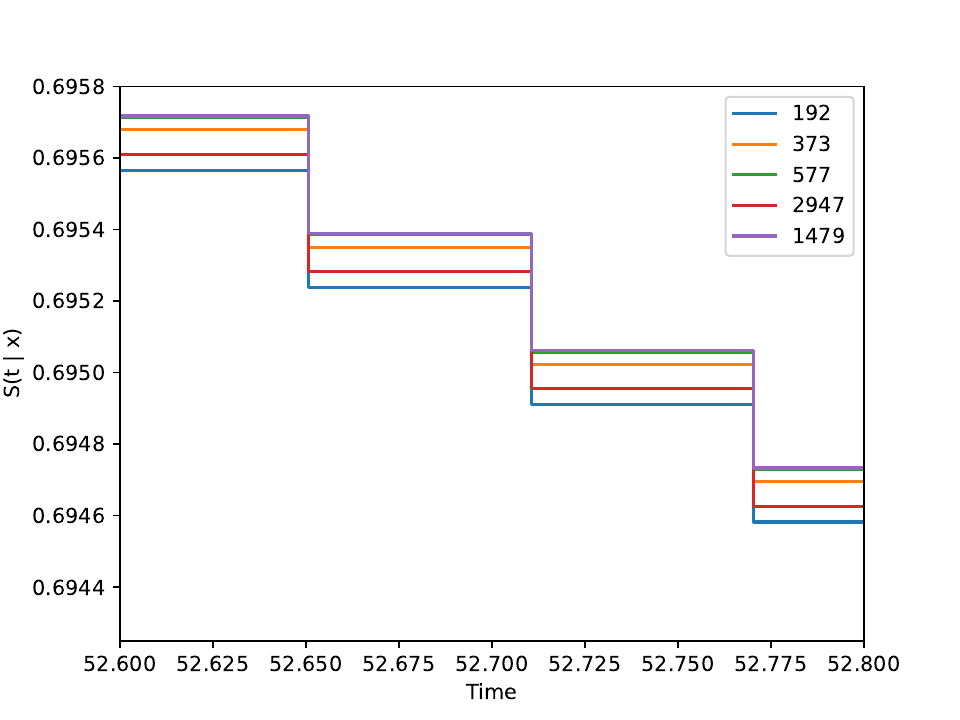}
\caption{DeepHit zoomed-in}
\label{figure1d}
\end{subfigure}

\caption{ Predicted OS curves for the same random set of five patients by three models respectively. The legend shows the patient ID. 
}
\label{figure1}
\end{figure}

Figure \ref{figure1} shows the predicted survival curves by three models of the same five patients.

%
%
%

\end{document}